\documentclass[11pt]{article}
\usepackage{eamt23}
\usepackage{times}
\usepackage{url}
\usepackage{latexsym}
\usepackage[small,bf]{caption} 
\usepackage{xcolor}
\usepackage{soul}
\usepackage{amsmath}

\title{Incorporating Human Translator Style into English-Turkish Literary Machine Translation}

\author{Zeynep Yirmibeşoğlu, Olgun Dursun, Harun Dallı,\\
\textbf{Mehmet Şahin, Ena Hodzik, Sabri Gürses, Tunga Güngör}\\
  Boğaziçi University\\
  Istanbul, Türkiye, 34342 \\
 {\tt \{zeynep.yirmibesoglu, olgun.dursun, harun.dalli, mehmet.sahin5,}\\
  {\tt  ena.hodzik, sabri.gurses, gungort\} @boun.edu.tr} }

\date{}

\begin{document}
\maketitle
\begin{abstract}

Although machine translation systems are mostly designed to serve in the general domain, there is a growing tendency to adapt these systems to other domains like literary translation. In this paper, we focus on English-Turkish literary translation and develop machine translation models that take into account the stylistic features of translators. We fine-tune a pre-trained machine translation model by the manually-aligned works of a particular translator. We make a detailed analysis of the effects of manual and automatic alignments, data augmentation methods, and corpus size on the translations. We propose an approach based on stylistic features to evaluate the style of a translator in the output translations. We show that the human translator style can be highly recreated in the target machine translations by adapting the models to the style of the translator.

\end{abstract}

\section{Introduction} 

Machine translation (MT) work has included literary texts in its agenda in the last decade and recent studies have shown some evidence for the possible contribution of machine translation in literary translation \cite{toral2015machine,Toral2018}. A few studies focused on the translator style in relation to machine translation (e.g., \newcite{kenny2020machine}), but to the best of our knowledge no research has embarked on building customized machine translation models evaluated on style metrics in literary texts.

In this paper, we aim at creating machine translation models which could generate outputs with literary style, particularly with the style of a translator. As a case study, we focus on the English-Turkish language pair. We make an analysis of literary style by following a hybrid methodology and identify the lexical and syntactic features that can reflect the translator’s style. We compile and manually align a corpus comprised of the works of a literary translator. By fine-tuning a pre-trained machine translation model on the corpus, we analyze in depth the effects of manual and automatic alignments, data augmentation techniques, and corpus size on both the translation quality and the style of the translations. We show that a machine translation system can be adapted to the style of a translator to obtain literary translations with that particular style.

The contributions of this paper to the field are as follows:
\begin{itemize}
    \itemsep0em 
    \item We introduce the first study in Turkish literary machine translation that trains models specific to a translator’s works
    \item We make a detailed analysis of literary style by following a hybrid methodology
    \item We build a manually-aligned corpus of a distinguished Turkish literary translator
    \item We devise a method that filters the alignments made by automatic alignment tools
    \item We make an in-depth analysis of translation quality and translator style in the literary domain
\end{itemize}

\section{Related Works}

\subsection{Style Analysis}

The concept of translator style has garnered growing interest in corpus-based translation studies. Some scholars maintain that stylistic traits can be observable by solely examining the target text \cite{baker2000towards}, whereas others inspect the target with consideration of the original author’s style \cite{malmkjaer2003happened,munday2008relations,saldanha2011translator,saldanha2014style}. Regardless of the influence of authorial style, the existence of translator style is unequivocal and its characteristics can be investigated independently, i.e., irrespective of authorial style and/or source text. The present study conceptualizes “translator style” as a consistent configuration of distinctive characteristics that are identifiable across multiple translations, and which exhibit a discernible impetus that is not explicable solely in terms of authorial style or linguistic limitations \cite{saldanha2011translator}.

In translation studies, corpus tools are used to observe patterns of stylistic choices based on comparisons between translation and reference corpora, the former representative of a particular translator and the latter of more general linguistic trends (see Baker's 2000 methodology). For example, type-token ratio (i.e., the ratio of the number of distinct words (types) to the total number of words (tokens), morpheme, word and sentence lengths, and frequency of lexical categories are considered indicators of vocabulary richness and lexical and syntactic complexity \cite{baker2000towards,li2011translation,saldanha2011translator}. Keyness analysis revealing not only frequent but also rare and specialized vocabulary of a translator \cite{olohan2004introducing} has also been used to compare between stylistic characteristics of human and machine generated translations. Important differences have been observed in lexical consistency between human and machine translations, in the sense that human translations have been found to be more explicit and target-oriented for the purpose of achieving better comprehension among their readers \cite{frankenberg2022can}.

\subsection{Literary Machine Translation}

Previous research in using machine translation in literary domain includes a variety of approaches for training and evaluation of the machine translation systems. \newcite{sluyter-gathje_neural_nodate} use both literary and out-of-domain data for English-German language pair with both statistical and neural methods. Their findings point towards statistical machine translation systems trained only with literary data being superior to other neural machine translation setup, and state the lack of large volume of literary data as a bottleneck.

\newcite{toral2015machine} explore the feasibility of using statistical machine translation (SMT) to translate a novel by Carlos Ruiz Zafon from Spanish into Catalan and they reach to the conclusion that literary MT is in its infancy. \newcite{Toral2018} show that neural machine translation models systematically outperform statistical models, especially with large datasets. These works do not focus on style of specific translators, but rather on generic literary machine translation.

\newcite{michel-neubig-2018-extreme} and \newcite{wang-etal-2021-towards} use a dataset of TED talks to replicate the translator style, the former using LSTMs and the latter using transformers. Both show promising results on the possibility of MT systems to capture translator style. \newcite{kuzman-etal-2019-neural} and \newcite{matusov-2019-challenges} employ fine-tuning of general purpose MT systems to capture literary style. \newcite{wang2022controlling} make use of style activation prompts to generate translations in the desired style, and propose a new benchmark called the multiway stylized machine translation (MSMT) benchmark. 

There are few studies involving the use of MT for literary texts in the English–Turkish language pair. \newcite{sahin2014translation} investigated the use of Google Translate\footnote{translate.google.com} (GT), which was using the SMT paradigm at the time, by novice translators for different text genres, including literary texts. \newcite{sahin-gurses-2019-mt} used GT after its switch to the NMT paradigm to analyze how it affected novice translators' creativity in literary retranslations. Based on qualitative analyses of their data and the results, the former study concluded that MT is unhelpful in literary translation, and the latter provided evidence that the use of MT has a restricting effect on novice translators' creativity.

\section{Corpus Compilation}

In this study, we have compiled two corpora, the translator corpus and the reference corpus \footnote{Copyright permissions for the usage of the books in the scope of this research have been taken. These permissions disallow us from making the corpora public.}. The translator corpus is an English-Turkish bilingual corpus and the reference corpus consists of Turkish monolingual texts. 

\subsection{Translator Corpus}
\label{TranslatorCorpus}

The translator corpus consists of the works of the literary translator Nihal Ye\u{g}inobalı  (1927-2020). As a distinguished literary translator, Ye\u{g}inobalı offers a fascinating case study for investigating translator style. During the last years of her career, she focused on writing her own literary works and also declared that she had published two pseudo-translations in the past years. Based on this we may believe that with the intention of being an author herself, Ye\u{g}inobalı  had incorporated idiosyncratic and personal elements to her translations that do not necessarily originate from the source text.

Between 1946 and 2013, Ye\u{g}inobalı produced a total of 129 works; she translated 123 books and authored six literary works of her own. The Yeğinobalı translator corpus has been digitized with the informed consent of her heirs in compliance with pertinent copyright laws. The digitization process entailed obtaining physical copies of the texts for scanning, refining the optically-read digital versions, and manually aligning the target texts with their corresponding source texts to train the machine translation models. Given the practical inaccessibility of certain texts, the digitized corpus comprises 100 optically-recognized texts, of which 56 were manually aligned. A total of 47 annotators worked on the manual alignment of the texts within the scope of this study. The experiments in this study were conducted with a sample of 51 manually aligned texts (48 for training and 3 for testing), as five texts were still in progress.

The manually-aligned 51 books contained many non-standard punctuations, which negatively affect the MT experiments. Thus, we normalized all hyphens, quotation marks, and apostrophes in the texts. Afterward, sentences have been tokenized with the SentencePiece \cite{kudo-richardson-2018-sentencepiece} tokenizer of the used Huggingface model \cite{huggingface_2019}. 

\subsection{Reference Corpus}
\label{ReferenceCorpus}

The stylistic investigation of a translator's style also involves a reference corpus, which serves to authenticate the idiosyncrasies by measuring them against accepted benchmark values. The reference corpus comprises 512 e-books, which are reflective of the linguistic tendencies that were prevalent in Turkish literary translations throughout Ye\u{g}inobalı's active period, from 1946 to 2013.  

\section{Translator Style Analysis}

\subsection{Methodology}

Drawing on Youdale’s \cite{youdale2022use} hybrid methodology, this study incorporates close and distant-reading techniques to counterbalance researcher bias in qualitative analysis and decontextualization of style in quantitative analysis. Close-reading is based on the checklist of style markers compiled by \newcite{leech2007style}, which comprises four levels of qualitative stylistic assessment: lexical, grammatical, semantic, and discourse. Distant-reading involves quantitative analysis of lexical and morphological stylistic traits, including a comparison of the translator corpus with a reference corpus to identify keywords and key clusters at the lexical level, and analysis of morphemes per sentence and word, including characteristic inflectional morphemes, at the morphemic level. Quantitative stylistic features are computed by means of average normalized frequency to ensure the comparability of results across texts of varying lengths. These traits are then contrasted with reference values to validate idiosyncrasies. In this work, we are focusing on the stylistic features of Nihal Ye\u{g}inobalı and the possibility of replicating her style in machine translation models. 

\subsection{Features}

Table \ref{table:style_features} displays the stylistic features and their categories used in this work. Through a combination of close and distant-reading sessions, we have identified a multitude of idiosyncratic lexical features that exhibit higher incidence rates in the translator corpus (Section~\ref{TranslatorCorpus}) compared to the reference corpus (Section~\ref{ReferenceCorpus}). Notable among these traits are the orthographic variant “gene” for the adverb “yine” \emph{(again)}, the conjunction “ki,”\footnote{Generally used as a translation of "that", "since", or "because".} and the conjunction cluster “gelgel+”\footnote{Literally, reduplication of "come". Generally used as a translation of "however", "nevertheless", or "still".} which comprises “gelgelelim” and “gelgeldim.” 

An equally intriguing lexical feature is the lower frequency of the conjunction “ve” (\emph{and}) compared to the reference value. This observation partially accounts for the heightened prevalence of alternative conjunctions in the translator corpus, indicating a propensity to avoid “ve” (\emph{and}).

\begin{table*}
\caption{Stylistic features used in translator style analysis}
\begin{center}
\resizebox{\textwidth}{!}{%
\begin{tabular}{|l|l|l|l|}
\hline
\multicolumn{1}{|c|}{\textbf{Word Level Features}} & \multicolumn{1}{c|}{\textbf{Sentence Level Features}} & \multicolumn{1}{c|}{\textbf{Morphological Data}} & \multicolumn{1}{c|}{\textbf{Focus Words}} \\ \hline
Type-token ratio & Ellipsis sentences & Average morphemes per sentence & "gelgelelim" \\
Number of unique words & Question sentences & Median morphemes per sentence & "gelgeldim" \\
Number of unique words, threshold = 10 & Exclamation sentences & Average morphemes per word & "maamafih" \\
Mean word length (characters) & Mean sentence length & Median morphemes per word & "gene", "ki", "ve" \\
Standard deviation of word lengths & Standard deviation of sentence lengths &  & "pek", "hem" \\
Reduplications & Median of sentence lengths &  &  "derken", "acaba "\\
 & Mode of sentence lengths &  & "sahiden" \\
 &  &  & "doğallıkla" \\ \hline
\end{tabular}
}
\label{table:style_features}
\end{center}
\end{table*}

\section{Automatic Alignment}
\label{AutoAlignment}

Manual alignment is a time-consuming job that requires meticulousness. Although it is absolutely necessary to manually align the English and Turkish books at least for the purpose of evaluation to arrive at reliable assessments, automatic alignment is a preferable method regarding human resources and time during the training phases. In this research, we worked with the \textit{hunalign} sentence aligner\footnote{https://github.com/danielvarga/hunalign} \cite{hunalign_2007} to automatically align the texts. However, the automatic alignment resulted in a considerable amount of erroneously aligned sentences, which deteriorated the translation performance when used as a parallel corpus. The problem was mostly caused by the omissions performed by the translator at hand from the original English text, or the merges of multiple English sentences into a single Turkish sentence. 

To eliminate the incorrectly aligned sentence pairs, we devised a method that makes use of machine translations of source sentences. The English sentence in each English-Turkish sentence pair in the \textit{hunalign} output is translated into Turkish using the pre-trained MT model that we use in this work (\textit{opus-mt-tc-big-en-tr}, see Section~\ref{MTModel}). By taking this translation as reference and the Turkish sentence in the \textit{hunalign} output as prediction, we computed the BLEU, METEOR, Google BLEU (GLEU, \newcite{wu2016googles}), and BERTScore F1 \cite{bertscore_2019} scores that evaluate the match between the two Turkish sentences. Taking these four scores as features, we trained an SVM \cite{Cortes1995SupportVectorN} model that predicts whether it is a correct alignment or not with a training set of 20 manually aligned books through the \textit{scikit-learn} library \cite{scikit-learn}. In all of our automatically aligned datasets explained in Section~\ref{Datasets}, we used this SVM model to extract the correct alignments from the \textit{hunalign} outputs and ignored the rest. 

\section{Machine Translation Model} \label{MTModel}

The Transformer architecture \cite{vaswani_2017} is dominant in the machine translation area, reaching state-of-the-art results in many language pairs. However, it is difficult to achieve high generalization in non-general domains, especially in the literary domain without a large training set. This is especially so if the research relies on capturing the style of a specific translator, in which case we face with the scarcity of the training data in addition to the cost and effort required in compiling and aligning the data. Even though all of the books of a translator are retrieved and aligned, the number of sentences may be as low as 200K. This amount of data is not adequate to train a successful Transformer model without augmentation. Taking Turkish-English machine translation at hand, the findings of WMT17 and WMT18 \cite{bojar-etal-2017-findings,bojar-etal-2018-findings} show that all of the participating systems make use of back-translation in some way or another, and the state-of-the-art results are achieved by The University of Edinburgh, where the initial news corpus of 200K sentences has been oversampled five times and augmented with 2.5M back-translated and 1M copied sentences \cite{haddow-etal-2018-university}. 

Keeping the importance of data in mind, we also observe a trend in NLP, where large pre-trained Transformer language models receive high popularity due to their success in various downstream tasks just by fine-tuning with much smaller training sets. The newest advances include text-to-text Transformer models such as T5 \cite{raffel_2019}, faster and more efficient ways of scaling and training text-to-text language models \cite{t5x_2022}, and combinations of different denoising objectives \cite{ul2_2022}. 

These recent trends brought to mind leveraging a large pre-trained machine translation model, and fine-tuning on the small training set that we obtain from the books of a specific translator. With this motivation, we selected Helsinki-NLP's English-Turkish pre-trained Transformer models trained as part of the OPUS-MT project\footnote{https://github.com/Helsinki-NLP/Opus-MT} \cite{tiedemann-thottingal-2020-opus}. The models have been trained on the English-Turkish OPUS corpus\footnote{https://opus.nlpl.eu/} \cite{tiedemann-2012-parallel} and the corpus gathered in the scope of the Tatoeba challenge \cite{tiedemann-2020-tatoeba} in the Marian-NMT framework \cite{junczys-dowmunt-etal-2018-marian}. We used the OPUS models in the Huggingface platform \cite{huggingface_2019}, specifically the \textit{opus-mt-tc-big-en-tr}\footnote{https://huggingface.co/Helsinki-NLP/opus-mt-tc-big-en-tr} model for the English-Turkish direction, which is the main translation direction in this research, since we aim to mimic the style of a Turkish translator. The Turkish-English translation direction has only been used for back-translation, where the \textit{opus-mt-tc-big-tr-en}\footnote{https://huggingface.co/Helsinki-NLP/opus-mt-tc-big-tr-en} model has been exploited. 

The English-Turkish pre-trained Transformer model has been fine-tuned on different training sets for 5 epochs which was seen as the optimal epoch number on the validation set, with a batch size of 64 fit into 4 Tesla V100 GPUs. The maximum source and target sentence lengths have been selected as 128, and the learning rate as \textit{2e-5} using the Adam optimizer with weight decay (0.1) \cite{adam_2017}.

\begin{table*}
\caption{Training Dataset Statistics}
\begin{center}
\scalebox{0.95}{
\begin{tabular}{|c|c|c|c|c|c|}
\hline
& \multicolumn{2}{c|}{\textbf{Manual}} & \multicolumn{2}{c|}{\textbf{Automatic}} & \textbf{Synthetic} \\ \hline
 & \textbf{Sentences} & \textbf{Books} & \textbf{Sentences} & \textbf{Books} & \textbf{Sentences} \\ \hline
   \textbf{Manual} & 283,810 & 48 & - & - & -\\ \hline
   \textbf{Manual-auto} & 121,009 & 24 & 120,834 & 24 & -\\ \hline
   \textbf{Auto} & - & - & 231,986 & 48 & -\\ \hline
   \textbf{Self-trained-small} & 283,810 & 48 & - & - & 250,000 \\ \hline
   \textbf{Self-trained-large} & 283,810 & 48 & - & - & 800,000 \\ \hline
   \textbf{Back-translated-small} & 283,810 & 48 & - & - & 250,000\\ \hline
   \textbf{Back-translated-large} & 283,810 & 48 & - & - & 800,000\\ \hline
\end{tabular}}
\label{table:train_dataset}
\end{center}
\end{table*}

\section{Augmentation}

Creating parallel data for training machine translation models is extremely challenging, whereas monolingual data in nearly all the languages are abundant. Literary machine translation requires a large amount of literary parallel data, which is currently unavailable and very expensive to align. Due to the low number of aligned literary data, two data augmentation methods have been carried out in this research to increase the quality of literary machine translation. 

\subsection{Back-translation}

\newcite{sennrich-etal-2016-improving} introduced back-translation, where automatic translation is performed on the monolingual data in the target side to generate synthetic sentences in the source side. This approach shows useful in many language pairs, reaching state-of-the-art results \cite{kocmi-etal-2022-findings}. 

Since the objective is to increase literary machine translation quality in the English-Turkish direction, first the Turkish-English OPUS-MT model has been fine-tuned on the 48 manually aligned books. This model has then been used to back-translate 800K randomly picked Turkish sentences (with minimum 3, maximum 128 tokens) obtained from 266 literary e-books to generate synthetic English sentences. The 800K parallel sentences have been coupled with the 48 manually aligned books. 

\subsection{Self-training}

We also experimented with self-training as a method of data augmentation. The difference is that the direction of augmentation is the same as the original translation direction, in that monolingual data from the source side is automatically translated into the target side. This way, monolingual English sentences are used to generate synthetic Turkish sentences. For this purpose, 800K sentences (with minimum 3, maximum 128 tokens) have been randomly picked from the English BookCorpus \cite{Zhu_2015_ICCV}. Since the BookCorpus contains only lowercase characters, the monolingual corpus has been truecased with the \textit{truecase} Python library. For automatic translation of the English sentences, we fine-tuned the English-Turkish OPUS-MT model on the 48 manually aligned books of the translator. Using this fine-tuned model, the 800K  sentences have been automatically translated into Turkish. 

\section{Stylistic Evaluation}

We quantify the style of a translation text using the set of 29 numeric features listed in Table \ref{table:style_features} and represent the text with a 29-dimensional vector \textbf{v} named as the \textit{style vector}. Since the features have different ranges and variances, we normalize the style vector \textbf{v} with min-max normalization:

\begin{equation*}
\hat{\textbf{v}}_i = \frac{\textbf{v}_i - \text{min}_i}{\text{max}_i - \text{min}_i} 
\end{equation*}

\noindent
where $i$ is the index of a feature, $\textbf{v}_i$ and $\hat{\textbf{v}}_i$ denote, respectively, the original value and the normalized value of feature $i$, and $\text{min}_i$ and $\text{max}_i$ denote, respectively, the minimum value and the maximum value of feature $i$ in the reference corpus. 

We use two metrics, cosine similarity and Pearson’s correlation coefficient, to measure the style match between the two translations of a text. The main motivation behind this choice is based on the assumption that texts with similar style have similar style vectors and these metrics adequately show the similarity between vectors. For the stylistic evaluation of a machine translation model on a test set, we take the translation output by the model and the original translation of the translator as the two translations and employ the similarity and correlation metrics on the style vectors. The expectation is to have high similarity and correlation scores if the model output is stylistically similar to the translation of the translator.

\section{Experiments and Results}

\subsection{Datasets}
\label{Datasets}

In order to observe the effect of manual and automatic alignments and data augmentation on the performance of the MT system and the style transfer, we built several training corpora of varying sizes. Table \ref{table:train_dataset} depicts the number of sentences and books and the alignment style for each corpus. The \textit{Manual} dataset consists of 48 manually aligned books from the translator corpus. \textit{Manual-auto} is a combination of 24 manually and 24 automatically aligned books, where the books were selected with a heuristic that balances the number of manually and automatically aligned sentences. The \textit{Auto} corpus consists of 48 automatically aligned books. We note that we obtained the automatically aligned books in the \textit{Manual-auto} and \textit{Auto} corpora by automatically aligning those books as explained in Section~\ref{AutoAlignment} rather than using their manual alignments.

In addition, the \textit{Manual} dataset has been augmented with self-training and back-translation. \textit{Self-trained-small} is a combination of \textit{Manual} and a portion of size 250K selected randomly from the 800K self-trained data. \textit{Self-trained-large} is formed in the same fashion and contains 800K synthetic parallel sentences. In a similar manner, \textit{Back-translated-small} consists of \textit{Manual} and a portion of size 250K sampled randomly from the 800K back-translated data. \textit{Back-translated-large} contains 800K back-translated sentences. The validation set is split randomly for each corpus and contains 5\% of the number of sentences in the training set. 

Similar to the training sets, we formed several test sets to observe the effects of the models on different types of data. Four test sets have been used for evaluation, two of which (\textit{Test-small} and \textit{Test-large}) contain manually aligned sentences. \textit{Test-large} is composed of the three manually aligned books (5,550 sentences) as a whole and is used both for quantitative evaluation and also for stylistic analysis. We noticed that the three books include very short or long sentences and may not be ideal for translation quality measurements. Therefore, by removing sentences with less than 4 and more than 25 tokens, we generated another test set (\textit{Test-small}) which contains 3,028 sentences. The other two test sets are benchmark news test sets from WMT17 (\textit{newstest2017}, \cite{bojar-etal-2017-findings}) and WMT18 (\textit{newstest2018}, \cite{bojar-etal-2018-findings}). 

\begin{table*}
\caption{Test set BLEU scores for the corpus size experiments. The best score for each test set is shown in bold.}
\begin{center}
\scalebox{0.95}{
\begin{tabular}{|c|c|c|c|c|}
\hline
  \textbf{Train Set}  & \textbf{Test-small} & \textbf{Test-large} & \textbf{newstest2017}  & \textbf{newstest2018} \\ \hline
   50K & 10.73 & 8.82 & \textbf{18.20} & \textbf{16.45} \\ \hline
    100K & 10.64 & 8.88 & 17.33 & 15.47  \\ \hline
    150K & \textbf{10.95} & 8.97 & 16.70 & 15.04  \\ \hline
    200K & 10.73 & 8.91 & 15.27 & 13.95 \\ \hline
    250K & 10.59 & 8.93 & 15.22 & 13.66  \\ \hline
    Manual (269K) & 10.89 & \textbf{9.04} & 15.02 & 13.27 \\ \hline
\end{tabular}}
\label{table:corpus_size_results}
\end{center}
\end{table*}

\begin{table*}
\caption{BLEU scores on the test sets, and cosine similarity (CS) and Pearson correlation coefficient (PC) results on \textit{Test-large} test set. The best score for each test set and style metric is shown in bold.}
\begin{center}
\scalebox{0.95}{
\begin{tabular}{|c|c|c|c|c|c|c|}
\hline
  \textbf{Train Set}  & \textbf{Test-small} & \textbf{Test-large} & \textbf{newstest2017}  & \textbf{newstest2018} & \textbf{CS} & \textbf{PC} \\ \hline 
  \textit{Pre-trained (Baseline)} & 7.23 & 5.81 & \textbf{25.47} & \textbf{22.58} & 0.681 & 0.408  \\ \hline 
   Manual & 10.89 & 9.04 & 15.02 & 13.27 & 0.923 & 0.807  \\ \hline
   Manual-auto & 10.61 & 8.80 & 15.59 & 13.89 & \textbf{0.952} & \textbf{0.886} \\ \hline
   Auto & 10.56 & 8.53 & 15.57 & 13.84 & 0.894 & 0.752  \\ \hline
   Self-trained-small & 10.69 & \textbf{9.05} & 13.51 & 12.30 & 0.856 & 0.645 \\ \hline
   Self-trained-large & 10.70 & 9.01 & 12.81 & 11.73 & 0.806 & 0.527  \\ \hline
   Back-translated-small & \textbf{10.94} & 8.88 & 18.39 & 16.17 & 0.885 & 0.715 \\ \hline
   Back-translated-large & 10.47 & 8.64 & 18.29 & 16.39 & 0.880 & 0.708 \\ \hline
\end{tabular}}
\label{table:results_all}
\end{center}
\end{table*}

\subsection{Impact of Corpus Size}

Manual alignment is an extremely time-consuming task that requires skilled annotators. The manual alignment of 48 books of the translator took months. This is not practical considering that the proposed style analysis framework may be employed for the works of several other translators later. Therefore, we conducted an experiment to analyze how many books or sentences could be adequate to both obtain a good translation quality and capture the translator's style. For this analysis, we obtained five different datasets of smaller sizes from the \textit{Manual} dataset having 50K, 100K, 150K, 200K, and 250K training sentences. Corresponding validation sets are 5\% of the training sets, as in other experiments.   

Table \ref{table:corpus_size_results} presents the results for the corpus size experiment, where the number of training sentences is shown for each model. Inference has been carried out on four test sets, for which the BLEU scores are provided to judge the translation quality of each model. The BLEU scores show a gradual improvement in literary translation quality when more literary training data is added. Interestingly, the news translation performance is compromised while the literary translation performance improves. As the models adapt more to the literary domain, the translations of news sentences get less accurate. The model with the highest BLEU score (10.95) for \textit{Test-small} is \textit{150K}, while the best BLEU score (9.04) for \textit{Test-large} was obtained from \textit{Manual}  (269K training sentences). It can be suggested that around 150K-200K sentences could be enough to obtain a good literary translation, and could be followed as a guideline during the compilation of future translators' works. 

\subsection{Results}

The English-Turkish OPUS-MT model has been fine-tuned on the training corpora for 5 epochs. The BLEU scores on the four test sets and the cosine similarity and Pearson correlation scores on the \textit{Test-large} set are shown in Table \ref{table:results_all}. We compare the models to the pre-trained OPUS-MT model that we accept as the baseline.

Fine-tuning on a literary training set immediately shows its positive effect on the literary test sets and its negative effect on the news test sets. After fine-tuning the pre-trained model with the \textit{Manual} dataset, we see 3.66 and 3.23 BLEU score improvements on the \textit{Test-small} and \textit{Test-large} sets, respectively. However, the translation performance drops drastically for both news test sets. We observe that literary translation and news translation do not go hand in hand. 

Automatic alignment success is also extremely important for current and future literary MT research due to the need of lightening the burden of manual alignment. The BLEU scores indicate that half manual, half automatic alignment decreases literary translation quality by 0.2-0.3 BLEU scores with respect to fully manual alignment. Besides, we observe a 0.3-0.5 BLEU score drop with fully automatic alignment. These are promising results since we still obtain much better literary translation than the pre-trained model, which was pre-trained on more than 108 million sentences from many different domains. This shows that \textit{hunalign} coupled with our automatic alignment filtering algorithm can be preferred for aligning new literary corpora, resulting in much faster alignment and much more parallel data than is possible with manual alignment. 

Models trained with augmented data yield the best scores for \textit{Test-small} and \textit{Test-large}. We observe that self-trained data augmentation (\textit{Self-trained-small}) outperforms other models in \textit{Test-large}, and back-translated data augmentation reaches the best performance in \textit{Test-small} and also improves news translation quality. We notice a 45-52\% improvement in \textit{Test-small} and a 47-56\% improvement in \textit{Test-large} compared to the pre-trained model scores. On the other hand, the improvements over the authentic (manually or automatically aligned) datasets are not so large when the addition of synthetic data (250K or 800K sentences) is considered. In general, we observe that improving literary translation quality is not very straightforward and amplifying the training set does not directly increase the BLEU scores. 

The cosine similarity (CS) and Pearson correlation (PC) scores of the pre-trained model are quite low indicating that the translations output with this model cannot reflect the style of the translator well. The models fine-tuned with manually or automatically aligned data reflect the style much better, having the best results obtained with the \textit{Manual-auto} model. The scores drop after including synthetic data. This may be attributed to the fact that, although the authentic datasets include only the works of the translator, the synthetic datasets include large amounts of data not originated from the translator. In the end, we comment that we can capture the stylistic features of the translator (Nihal Yeğinobalı) much better than the pre-trained model when fine-tuned on her translations.

\section{Conclusions}

In this paper, we proposed an approach for literary machine translation that can adapt itself to the style of a translator and produce translations close to that style. As a case study, we focused on the English-Turkish language pair and a distinguished Turkish literary translator. In this direction, we leveraged a large pre-trained machine translation model and fine-tuned it on the works of the translator. The experiments were conducted using both manually and automatically aligned data compiled from the books of the translator. We also tested the effect of two data augmentation methods, self-training and back-translation, on the performance. To measure how much the translations obtained by the fine-tuned model reflect the style of the translator, we made a detailed analysis of literary style and identified a set of stylistic features. The experiments showed that adapting a pre-trained model to the works of a translator increases the BLEU score about 45-56\% on the literary data and captures the translator’s style 18-40\% better in terms of cosine similarity compared to the pre-trained model.

As future work, we plan to incorporate other evaluation metrics in addition to the BLEU score that can capture the semantics of the translations better. We also aim at conducting a human evaluation for both translation quality and stylistic properties. Another interesting direction will be including other literary translators, adapting the machine translation models to different styles, and experimenting with style transfer between works of the translators.

\section*{Acknowledgements}

This research is funded by the Scientific and Technological Research Council of Türkiye (TÜBİTAK) under Grant No: 121K221 (Literary Machine Translation to Produce Translations that Reflect Translators’ Style and Generate Retranslations). The numerical calculations reported in this paper were fully performed at TÜBİTAK ULAKBIM, High Performance and Grid Computing Center (TRUBA resources). 


\bibliography{eamt23}
\bibliographystyle{eamt23}
\end{document}